\documentclass[lettersize,journal]{IEEEtran}
\usepackage{amsmath,amsfonts}
\usepackage{algorithmic}
\usepackage{algorithm}
\usepackage{array}
\usepackage{textcomp}
\usepackage{stfloats}
\usepackage{url}
\usepackage{verbatim}
\usepackage{graphicx}
\usepackage{cite}
\usepackage{subfigure}
\usepackage{url}
\usepackage{multirow}
\usepackage{enumitem}
\setlist[enumerate]{leftmargin=0.9cm, label=(\arabic*)}
\usepackage{arydshln}
\usepackage{amsfonts}
\usepackage{amsmath}
\usepackage{amsthm}

\theoremstyle{definition}
\newtheorem{definition}{Definition}
\usepackage{booktabs}
\usepackage{xcolor}

\begin{document}

\title{Unsupervised Entity Alignment Based on Personalized Discriminative Rooted Tree}

\author{Yaming Yang, Zhe Wang, Ziyu Guan, Wei Zhao$^{*}$, Xinyan Huang, Xiaofei He
\thanks{* Corresponding author}
\thanks{Y. Yang, Z. Wang, Z. Guan, and W. Zhao are with the State Key Laboratory of Integrated Services Networks, School of Computer Science and Technology, Xidian University, Xi'an, China 710071. E-mail: \{yym@, zwang\_01@stu., zyguan@, ywzhao@mail.\}xidian.edu.cn}%
\thanks{X. Huang is the Key Laboratory of Intelligent Perception and Image Understanding of Ministry of Education, School of Artificial Intelligence, Xidian University, Xi'an, China 710071. E-mail: huangxy@xidian.edu.cn}%
\thanks{X. He is with the State Key Laboratory of CAD\&CG, Zhejiang University, Hangzhou, China. E-mail: xiaofeihe@cad.zju.edu.cn}%
}

\markboth{Journal of \LaTeX\ Class Files, February~2025}%
{Yang \MakeLowercase{\textit{et al.}}: Unsupervised Entity Alignment Based on Personalized Discriminative Rooted Tree}


\maketitle

\begin{abstract}
Entity Alignment (EA) is to link potential equivalent entities across different knowledge graphs (KGs). Most existing EA methods are supervised as they require the supervision of seed alignments, i.e., manually specified aligned entity pairs. Very recently, several EA studies have made some attempts to get rid of seed alignments. Despite achieving preliminary progress, they still suffer two limitations: (1) The entity embeddings produced by their GNN-like encoders lack personalization since some of the aggregation subpaths are shared between different entities. (2) They cannot fully alleviate the distribution distortion issue between candidate KGs due to the absence of the supervised signal. In this work, we propose a novel unsupervised entity alignment approach called UNEA to address the above two issues. First, we parametrically sample a tree neighborhood rooted at each entity, and accordingly develop a tree attention aggregation mechanism to extract a personalized embedding for each entity. Second, we introduce an auxiliary task of maximizing the mutual information between the input and the output of the KG encoder, to regularize the model and prevent the distribution distortion. Extensive experiments show that our UNEA achieves a new state-of-the-art for the unsupervised EA task, and can even outperform many existing supervised EA baselines.
\end{abstract}


\section{Introduction}
\label{sec:introduction}
\IEEEPARstart{K}nowledge Graph (KG) can describe massive knowledge facts of our world and has been successfully applied to various tasks, such as question answering~\cite{answer}, search engines~\cite{zhao2021brain}, document retrieval~\cite{search}, recommender systems~\cite{recommdation}, etc. However, as a KG usually covers only a small part of knowledge facts of a specific domain, it may fail to provide sufficient knowledge to support downstream applications. Entity Alignment (EA) is an effective solution for this issue. It can fuse multiple KGs by identifying equivalent entities across KGs, and the merged KGs can provide more comprehensive information for downstream tasks. The primary challenge for EA comes from the heterogeneous symbolic representations of different KGs, which include different naming rules and multilingualism.

Existing EA methods~\cite{zeng2021comprehensive, zhao2020experimental, sun13benchmarking, zhang2022benchmark} address the heterogeneity issue by projecting different KGs into a common low-dimensional embedding space, where similar entities are pulled close while dissimilar ones are pushed far away. Thus, the similarity between entities can be conveniently measured by various distance functions such as cosine distance, and $l_1 / l_2$ norm. Based on how the entity embeddings are extracted, existing EA methods can be divided into two categories. Trans-based~\cite{mtranse, jape, transedge} methods utilize translation-family embedding methods, such as TransE~\cite{transe}, to learn entity representations. GNN-based EA methods~\cite{gcn-align, alinet, rdgcn, avrgcn} utilize Graph Neural Networks (GNNs)~\cite{gcn, gat} to encode KGs. Unfortunately, the majority of the existing methods, whether Trans-based or GNN-based, are supervised and their effectiveness relies on the supervision of high-quality manual labels, i.e., seed alignments. This is impractical since in the real world, it is expensive to obtain enough high-quality labels, and sometimes labels are unavailable due to privacy concerns. 

Recently, researchers have proposed several unsupervised EA methods~\cite{eva, iclea, uplr, selfkg, dualmatch} to make some efforts in getting rid of seed alignments. They typically use graph attention-based KG encoders~\cite{gat} to extract entity embeddings. Based on the distance of these entity embeddings, they generate pseudo-labels to serve as the self-supervised alignment signal. Despite showing preliminary progress, they still face two limitations.

\begin{figure}[ht]
\centering
\subfigure[``Bill Gates'' prefers second-order neighbor ``Steve Ballmer'', while ``Private School'' prefers ``Ivy League'', creating a dilemma for the common neighbor ``Harvard University''.]{\includegraphics[width=1\columnwidth]{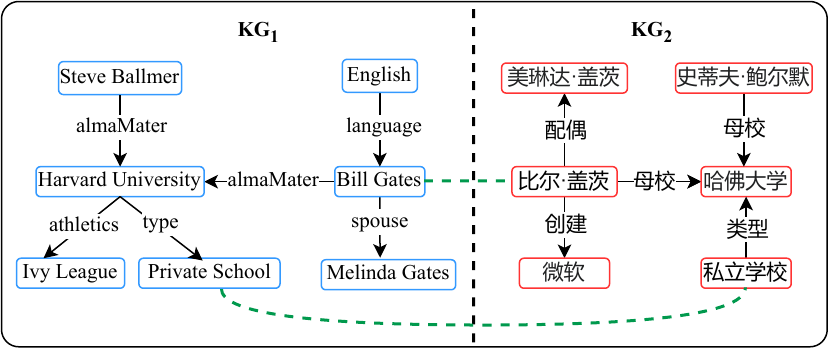}\label{fig:exam_1}}\\
\subfigure[In the unsupervised setting, the embedding distribution could be distorted due to the wrong guidance of false pseudo-labels.]{\includegraphics[width=1\columnwidth]{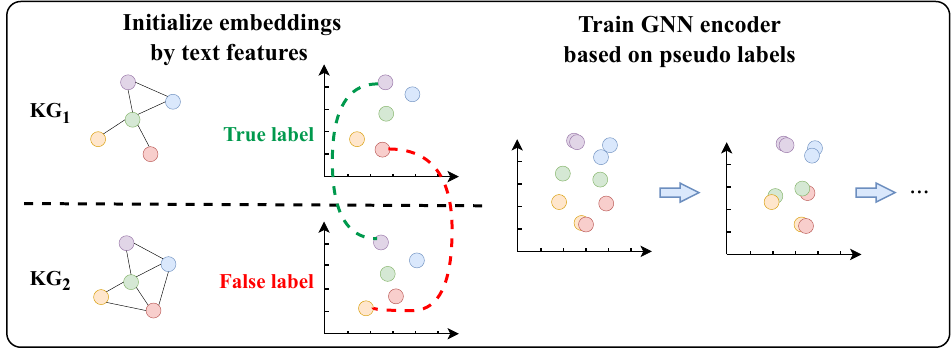}\label{fig:exam_2}}  
\caption{Illustration of the two limitations of existing EA methods.}
\label{fig:vis}
\end{figure}

\textbf{Limitation 1: low personalization of entity embeddings.}
Current unsupervised EA methods generally stack multiple GNN layers and extract entity embeddings through iterative attention aggregation. However, this encoding scheme may not be able to extract the most discriminative embedding for an entity to find its potential alignment counterpart. This is because some aggregation subpaths are inevitably shared between different entities, limiting the personalization and flexibility of entity embeddings. 

\begin{figure}[ht]
\centering
\includegraphics[width=1\columnwidth]{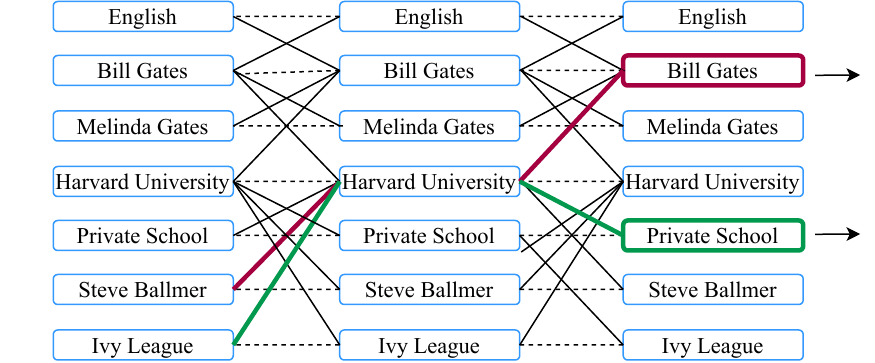}
\caption{Visualization of GNN's aggregation paths for the six entities that are illustrated in Figure 1(a).}
\label{fig:path-ent}
\end{figure}

For example, Fig.~\ref{fig:exam_1} shows two toy KGs to be aligned. When the model tries to align the English entity ``Bill Gates'' of KG$_1$ with the Chinese entity ``Bill Gates'' of KG$_2$, their second-order neighbor ``Steve Ballmer'' is more discriminative. Contrastively, when examining the English entity ``Private School'' and its Chinese counterpart, their second-order neighbor ``Ivy League'' is more discriminative. To further analyze this issue, in Fig.~\ref{fig:path-ent}, we visualize the aggregation paths of a GNN for the six entities of KG$_1$ in Fig.~\ref{fig:exam_1}. Solid lines represent neighborhood aggregation, and dotted lines represent self-connections. We can see that the target entity ``Bill Gates'' wants its second-order neighbor ``Steve Ballmer'' to have a larger aggregation weight but wants another second-order neighbor ``Ivy League'' to have a smaller aggregation weight. However, the situation is the opposite for the target entity ``Private School''. This conflicting situation creates a dilemma for their common first-order neighbor ``Harvard University'' when it aggregates its own neighbors.

\textbf{Limitation 2: cannot fully avoid distribution distortion.}
Existing EA approaches can be generally abstracted into two key parts. Firstly, KG encoders are used to shape the original candidate KGs into a specific distribution in a low-dimensional Euclidean space. Secondly, based on the distribution, potentially aligned entity pairs are pulled together and unlikely pairs are pushed apart. The previous study RREA~\cite{rrea} proves that GNN-based EA methods are subject to this framework. Besides, it proves that the orthogonality of GNN's projection parameters is highly beneficial for the EA task since it corresponds to rotation transformation that does not affect the embedding distribution. However, this good property requires an important premise that the EA task is supervised. Besides, GNN's aggregation operator does not guarantee this property. Therefore, in the unsupervised setting, the embedding distribution could still be distorted due to the wrong guidance of false pseudo-labels. 

\begin{figure}[ht]
\centering
\subfigure[After LLM initialization and before the GNN aggregation.]{\includegraphics[width=1\columnwidth]{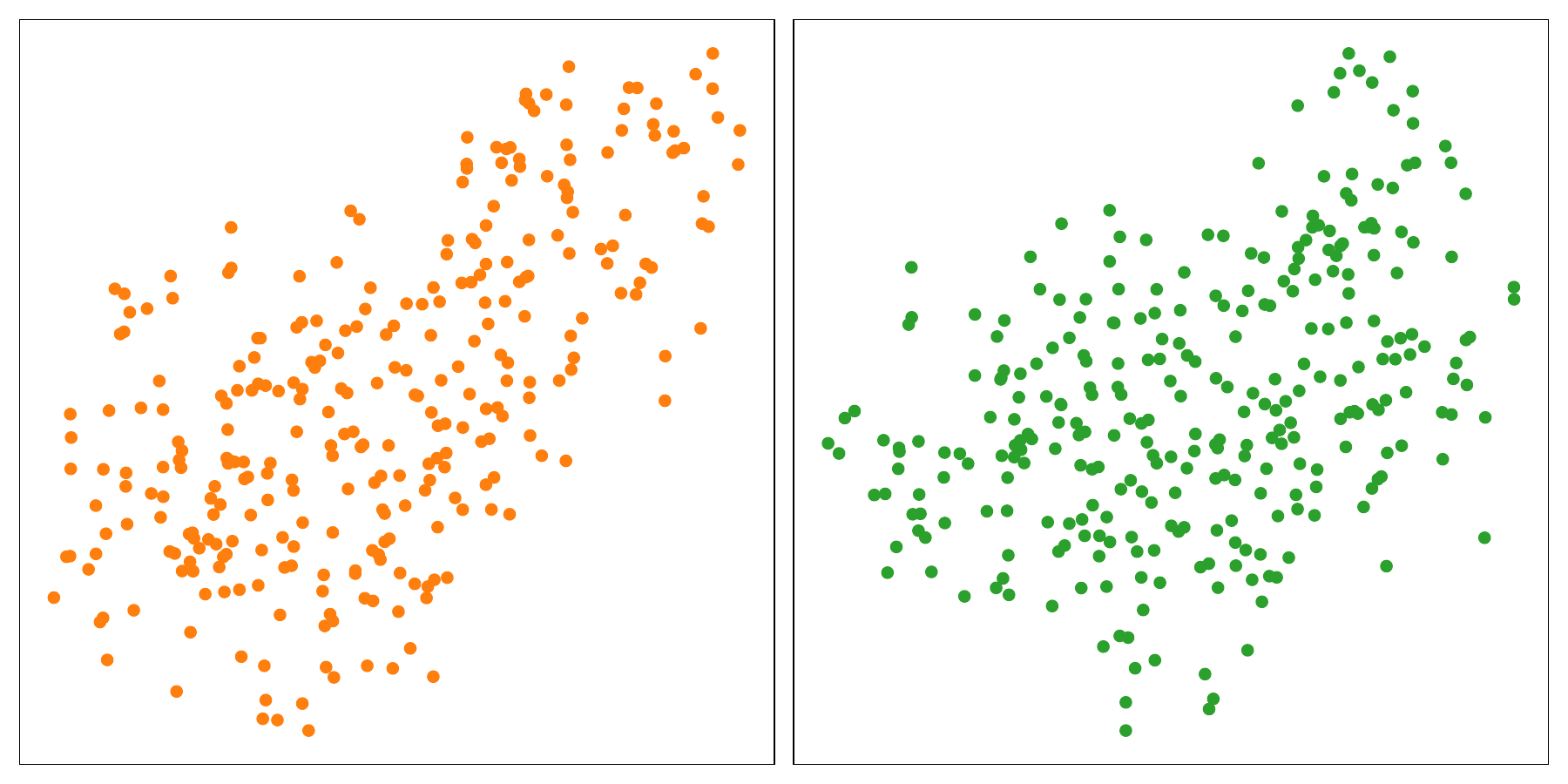}\label{fig:llm-init}}\\
\subfigure[After the GNN aggregation.]{\includegraphics[width=1\columnwidth]{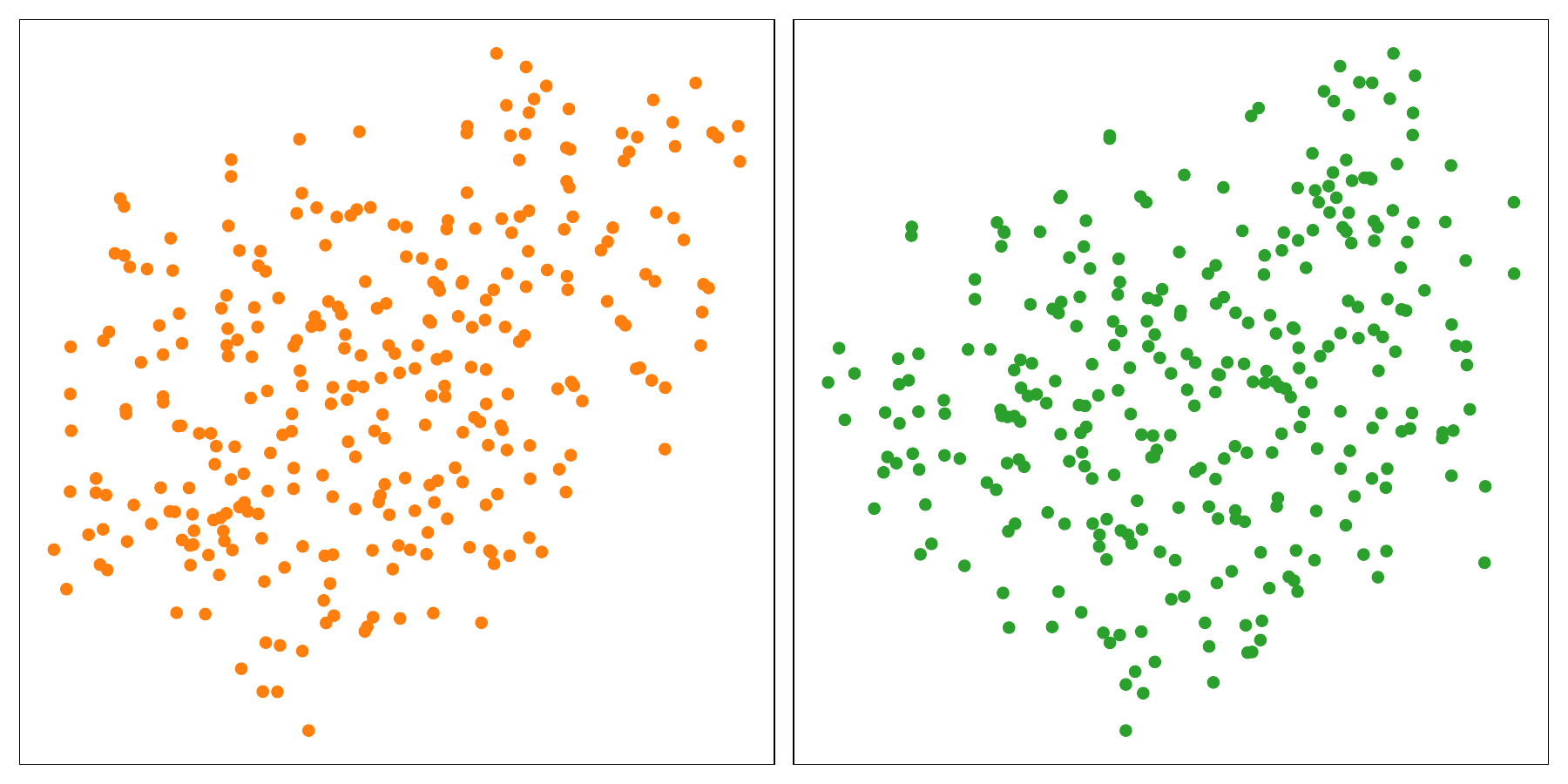}\label{fig:gnn-agg}}\\
\subfigure[After the GNN aggregation that incorporates our regularization terms.]{\includegraphics[width=1\columnwidth]{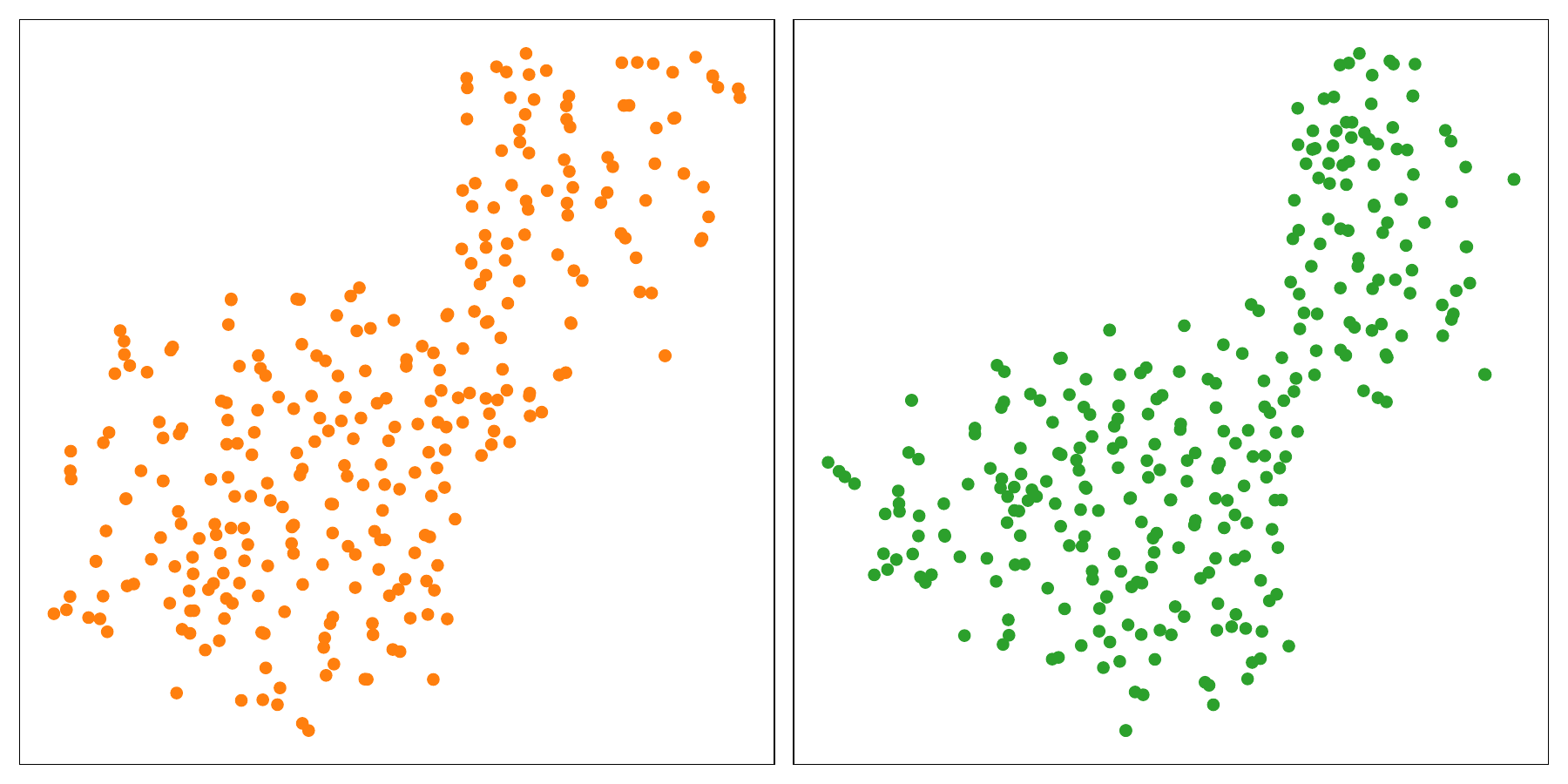}\label{fig:mireg}}
\caption{The entity embedding distributions of the two candidate KGs in the DBP15K$_{\text{fr\_en}}$ dataset under the unsupervised setting, with 300 pairs of entities from the French KG (left) and the English KG (right), respectively.}
\label{fig:visual-embd}
\end{figure}

As illustrated in Fig.~\ref{fig:exam_2}, the red dotted line mistakenly connects two nodes as an alignment pair. This error alignment signal may gradually accumulate, finally leading to distribution distortion and performance degeneration. To show this issue more intuitively, we randomly sample 300 pairs of entities from the two candidate KGs in the DBP15K$_{\text{fr\_en}}$ dataset (refer to Section~\ref{subsec:datasets} for more details) and visualize their embedding distributions. Specifically, we first utilize large language models (LLMs) to initialize entity embeddings based on their surface names and visualize them in Fig.~\ref{fig:llm-init}. Then, based on the initialized embeddings, we further conduct the GNN aggregation on them, and visualize the result in Fig.~\ref{fig:gnn-agg}. As we can see, the embedding distributions of the two KGs in Fig.~\ref{fig:llm-init} are more similar than those in Fig.~\ref{fig:gnn-agg}. This implies that, in our context, the GNN aggregation operation may not prevent the distribution distortion well.

In this paper, we propose a novel unsupervised EA method called UNEA to address the two limitations described above that hinder existing unsupervised EA methods.

Firstly, we use powerful LLMs to initialize the embeddings of entities and relations based on their surface names, which can be treated as a good ``weak supervision signal" for model optimization. As shown in Fig.~\ref{fig:llm-init}, after the entity embeddings are initialized by LLMs, the distributions of the two candidate KGs are already quite similar.

Secondly, we define a parametric sampling function to extract a discriminative tree neighborhood for each entity. In the sampled tree, each target entity itself serves as the root. Then, we design a corresponding tree attention aggregation mechanism to extract embedding for the target (root) entity. Since we customize a tree neighborhood for each target entity, their aggregation paths are fully decoupled. Thus, each target entity can learn its best aggregation path, resulting in more personalized embeddings.

Finally, to regularize the main alignment task, we let the model maximize the mutual information between the output high-level embeddings of entities and relations and their embeddings initialized by LLMs, thereby preserving the information of the ``weak supervision signal". Besides, we maximize the graphical mutual information~\cite{gmi} between the high-level entity embeddings and the KG topology. These mutual information maximization-based terms can continuously regularize the model so that the high-level embeddings of entities and relations can always reflect the information of their surface names and their topological relationships, thus preventing the possible issue of distribution distortion. In Fig.~\ref{fig:mireg}, we visualize the entity embedding distributions of two KGs that are extracted by the GNN that incorporates our regularization terms. As we can see, in comparison with Fig.~\ref{fig:llm-init} and Fig.~\ref{fig:gnn-agg}, the two embedding distributions in Fig.~\ref{fig:mireg} become tighter and are more similar to each other, indicating its effectiveness.

The main contributions of this work are summarized as follows:
\begin{itemize}
\item  We sample a personalized tree and propose an innovative tree attention aggregation mechanism to learn more personalized embeddings for entities.
\item We introduce mutual information maximization-based terms to continuously regularize the EA model, which helps avoid the distribution distortion issue in the unsupervised setting.
\item We conduct extensive experiments on two widely used benchmark datasets to verify the effectiveness of UNEA. It turns out that our UNEA can outperform state-of-the-art unsupervised baselines as well as supervised baselines, indicating its superiority.
\end{itemize}

\section{Related Work}
\label{sec:relate}
In this section, we review existing EA methods in terms of both supervised EA methods and unsupervised EA methods.

\subsection{Supervised Entity Alignment}
Most of the existing EA methods are supervised since they require seed alignment for supervision. According to their KG encoders, they can be divided into two categories~\cite{jiang2024toward, chen2022multi}. Trans-based methods~\cite{mtranse, jape, bootea, multike, jewp, jape, iptranse, transedge, neoea, mrpea, transedge} typically use translation-family KG encoders such as TransE~\cite{transe} to encode the structural information of KGs. GNN-based methods~\cite{gcn-align, mugnn, alinet, rdgcn, avrgcn, hgcn-je, rnm, rrea, tang2023weakly, liu2023dependency, li2023attribute, sun2022revisiting, hyperka, rea, epea, attrgnn, hman, hgcn-je-jr, gmnn, mugnn, nmn, ssp, alinet, mraea, naea, kecg, rrea, psr, largeea, roadea, cyctea, multiea,gala} use more advanced GNN-like encoders~\cite{gcn, gat}, usually in combination with relation-specific attention aggregation schemes~\cite{nathani2019learning, sheikh2021knowledge}to learn entity embeddings. After encoding, they perform alignment training by pulling seed alignment entity pairs closer while pushing the other entity pairs far away in the embedding space. There are also some methods that belong neither to Trans-based methods nor to GNN-based methods. For example, LightEA~\cite{lightea} achieves supervised alignment based on the label propagation algorithm. Although supervised EA methods have achieved remarkable alignment performance, they rely on manually provided high-quality supervised signals that are expensive to acquire in practice, hindering their widespread application in real-world scenarios.

\subsection{Unsupervised Entity Alignment}
Most existing unsupervised EA methods~\cite{eva, selfkg, uplr, dualmatch, iclea, emgcn} are GNN-based, and they usually utilize GCN~\cite{gcn} or GAT~\cite{gat} to extract the structural information of candidate KGs. Based on the extracted entity embeddings, they compute the similarity between all the possible entity pairs across candidate KGs. Then, they will select a set of the most similar entity pairs to form pseudo-labels to guide the alignment training. To improve the reliability of pseudo-labels, they often need to introduce various auxiliary information. For example, SelfKG~\cite{selfkg}, ICLEA~\cite{iclea}, and UPLR~\cite{uplr} introduce the text features of entity names, EVA~\cite{eva} introduces the image features associated with entities, and DualMatch~\cite{dualmatch} introduces the temporal information associated with relational facts. There is also a Trans-based unsupervised EA method, i.e., MultiKE~\cite{multike}, which leverages multiple views of entity features, such as names, relations, and attributes to enhance alignment. Despite showing preliminary progress in unsupervised EA, they still face the two limitations as we have analyzed in Section~\ref{sec:introduction}. In this work, we aim to develop a novel KG encoding technique and introduce an advanced mutual information-based regularization mechanism, to effectively address the two limitations and further boost the performance of the unsupervised EA task. 

There are also several unsupervised EA methods that belong neither to GNN-based methods nor to Trans-based methods. SEU~\cite{seu} and DATTI~\cite{datti} transform the EA problem into an assignment problem. ERAlign~\cite{eralign} jointly performs entity alignment and relation alignment by neighbor triple matching strategy, based on semantic textual features of entities and relations. FGWEA~\cite{fgwea} achieves the unsupervised EA task based on the Fused Gromov-Wasserstein Distance~\cite{fgw}. While effective, these methods often require complex optimization operations and sacrifice the ability of end-to-end learning.

\begin{figure*}[ht]
\centering
\includegraphics[width=2.0\columnwidth]{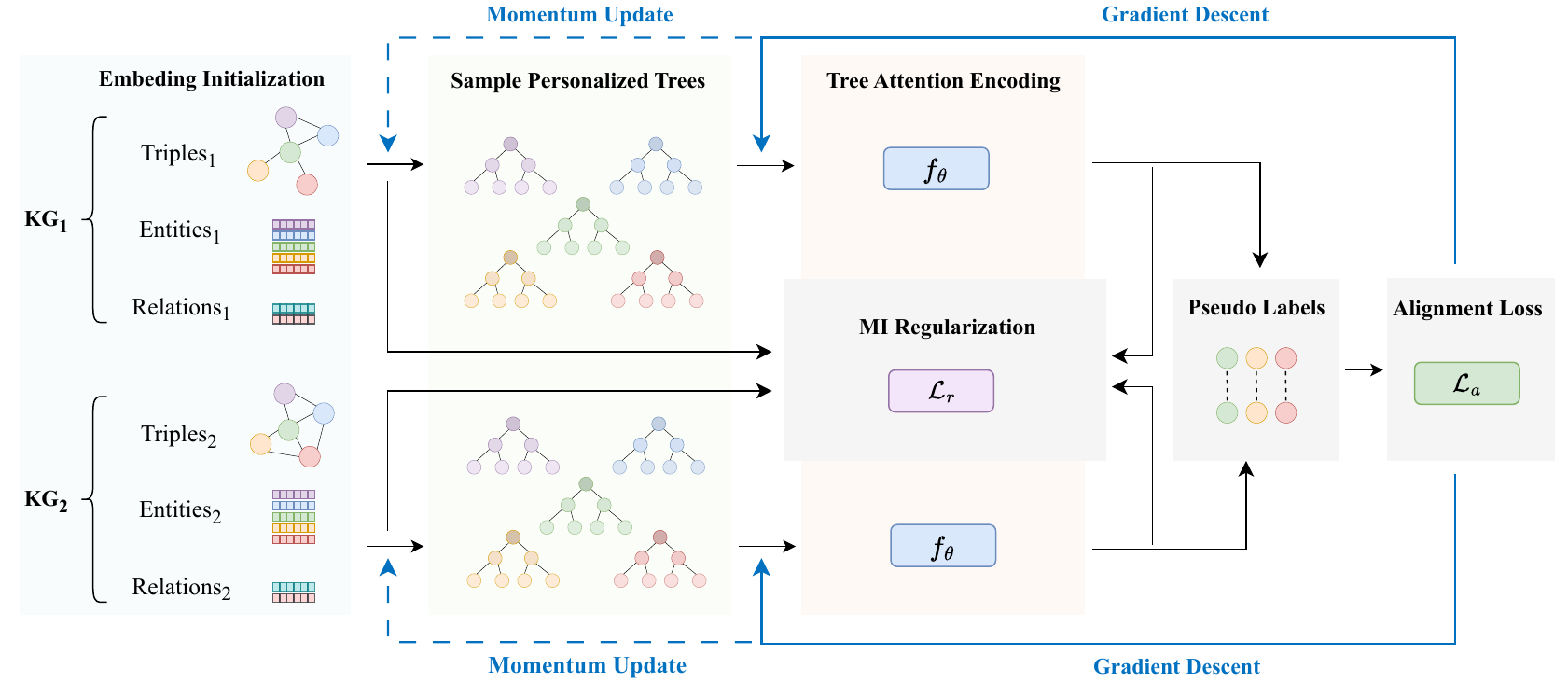}
\caption{The overall architecture of the proposed UNEA. The inputs include the triples, the entity embeddings, and the relation embeddings of two candidate KGs. The detail of each module is specifically described in Section~\ref{sec:method}.}
\label{fig:overview}
\end{figure*}

\section{Preliminaries}
\label{sec:pre}
In this section, we first define the basic notations that we use throughout the text, and then formally describe the problem that we study in this work.

\begin{definition}
\textbf{Knowledge Graphs.}
A knowledge graph $\mathcal{G}$ is represented as $\mathcal{G} = (\mathcal{E}, \mathcal{R}, \mathcal{T})$, where $\mathcal{E}$, $\mathcal{R}$, $\mathcal{T}$ denote the set of entities, the set of relations, and the set of triples, respectively. A triple $t \in \mathcal{T}$ can be denoted as $<i, k, j>$, depicting that the head entity $i \in \mathcal{E}$ and the tail entity $j \in \mathcal{E}$ hold the relation $k \in \mathcal{R}$ between them. Let $\mathbf{e}_i$ denote the representation vector of entity $i$, and $\mathbf{r}_k$ denote the representation vector of relation $k$.
\end{definition}

\begin{definition}
\textbf{Entity Alignment.}
Given a pair of candidate knowledge graphs $\mathcal{G}_1 = (\mathcal{E}_1, \mathcal{R}_1, \mathcal{T}_1)$ and $\mathcal{G}_2 = (\mathcal{E}_2, \mathcal{R}_2, \mathcal{T}_2)$ to be aligned, supervised entity alignment is to discover potential equivalent entity pairs, based on the supervision of a set of seed alignments $\mathcal{S} \subset \mathcal{E}_1 \times \mathcal{E}_2$, while unsupervised entity alignment aims to achieve this goal without any manual supervision signal.
\end{definition}

In this work, we concentrate on the more challenging setting of unsupervised entity alignment, where a set of pseudo alignments $\mathcal{S}$ are automatically generated and updated by the algorithm, without any manual labels.

\section{Methodology}
\label{sec:method}
In this section, we elaborate on the technical details of our proposed UNEA. Fig.~\ref{fig:overview} depicts its overall architecture.

\subsection{Embedding Initialization}
In recent years, LLMs have made significant strides across various domains. Many previous EA methods~\cite{hman,bert-int,selfkg,iclea} have utilized language models to initialize entity embeddings, which has substantially enhanced EA performance. In this work, we employ LLMs to initialize entity embeddings and relation embeddings. Specifically, we utilize LLMs to extract the name feature of entities and relations, which can be described as follows:
\begin{equation}
\label{eq:labse}
\begin{split}
\mathbf{f}_i^e &= \text{LLM}_{\theta}(n_i^e)\\
\mathbf{f}_k^r &=  \text{LLM}_{\theta}(n_k^r)
\end{split}
\end{equation}
where $\theta$ is the parameters of LLMs that are pre-trained in an unsupervised manner. The terms $n_i^e$ and $n_k^r$ are the surface names of entity $i$ and relation $k$, respectively. The vectors $\mathbf{f}_i^e$ and $\mathbf{f}_k^r$ are the corresponding features returned by LLMs, and we use them to initialize the entity embedding $\mathbf{e}_i$ and the relation embedding $\mathbf{r}_k$, as follows:
\begin{equation}
\label{eq:init}
\begin{split}
\mathbf{e}_i &= \text{Init}(\mathbf{f}_i^e)\\
\mathbf{r}_k &= \text{Init}(\mathbf{f}_k^r)
\end{split}
\end{equation}
During the training, $\mathbf{e}_i$ is updated by parametric GNN-like aggregation, and $\mathbf{r}_k$ itself is a learnable parameter vector.

\subsection{Sampling Personalized Rooted Trees}
\label{sbusec:sample-tree}

\begin{figure}[ht]
\centering
\includegraphics[width=1\columnwidth]{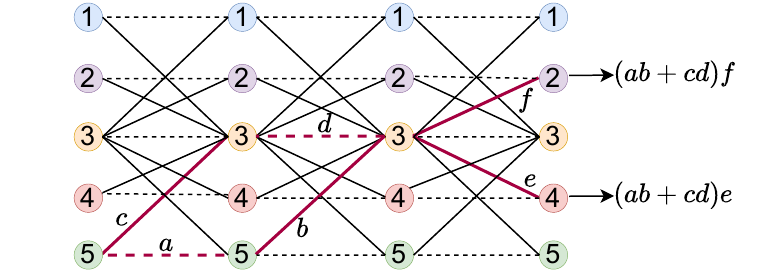}
\caption{Visualization of GNN's aggregation paths for five random entities. Some aggregation weights are marked on the paths.}
\label{fig:path-node}
\end{figure}

Recall that in Limitation 1 of Section~\ref{sec:introduction}, we have intuitively analyzed that previous EA-oriented GNNs have a limitation of coupling some sub-paths for aggregation. Here, we further explain this point in general. As shown in Fig.~\ref{fig:path-node}, we visualize the aggregation paths of a three-layer GNN model for five random entities. As we can see, a subset of the aggregation paths (red lines) from entity 5 to entity 2 is associated with the weight $(ab + cd)f$, and another subset of the aggregation paths from entity 5 to entity 4 is associated with the weight $(ab + cd)e$. As we can see, the subpaths associated with the weight $(ab + cd)$ are shared by entity 2 and entity 4, limiting the personality of their embeddings since they are obtained through this shared weighted aggregation.

To address this limitation, in this work, we propose to sample a discriminative neighborhood for each entity, and thus each entity can learn its own personalized aggregation paths for alignment.

\begin{figure}[ht]
\centering
\includegraphics[width=0.8\columnwidth]{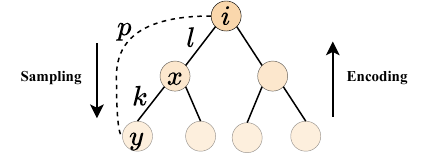}
\caption{Illustration of rooted tree sampling and encoding. Entity $i$ is the root of the tree. Entity $x$ is one of its first-order neighbors, and entity $y$ is one of its second-order neighbors. $l$ is the relation between $i$ and $x$, $k$ is the relation between $x$ and $y$, and $p$ is the composite relation between $i$ and $y$. The sampling process is from the root to leaves, and the encoding process is the opposite, i.e., from leaves to the root.}
\label{fig:path}
\end{figure}

As illustrated in Fig.~\ref{fig:path}, we take entity $i$ as the root and recursively sample its first-order neighbors, second-order neighbors, and so on. Suppose that we have sampled the first-order neighbor $x$ for root entity $i$, we further sample its second-order neighbor $y$ based on an attention distribution, described as follows:
\begin{equation}
\label{eq:sample}
\beta^{(i)}_{x,k,y} = \frac{ \sigma(\mathbf{e}_{i}^{T}  \cdot \mathbf{W}_{p} \cdot \mathbf{e}_{y} + \mathbf{e}_{x}^{T}  \cdot \mathbf{W}_{k} \cdot \mathbf{e}_{y}) } { log(d_y) }
\end{equation}
\begin{equation}
\label{eq:sample-att}
\alpha^{(i)}_{x,k,y} =\frac{ exp(\beta^{(i)}_{x,k,y}) }{ \sum\limits_{z \in \mathcal{N}_x, s = \phi(x,z)} exp(\beta^{(i)}_{x,s,z}) }
\end{equation}
where $\sigma (\cdot)$ is the non-linear activation function, $\mathcal{N}_x$ is a set denoting the neighbors of entity $x$, $s = \phi(x,z)$ denotes the relation $s$ between entities $x$ and $z$, and $d_y$ denotes the degree of entity $y$, which is used to reflect the fact that the discriminability of a neighbor is usually inversely proportional to its degree~\cite{nmn}. $\mathbf{W}_{k}$ is a parametric projection matrix specific to relation $k$. Following~\cite{rrea}, we set $\mathbf{W}_{k}$ as follows:
\begin{equation}
\label{eq:proj-w-k}
\mathbf{W}_{k} = \mathbf{I} - 2 \cdot \mathbf{r}_{k} \cdot \mathbf{r}_{k}^{T}
\end{equation}
This setting can naturally guarantee the orthogonality of the projection matrix since we can easily derive the following equation: 
\begin{equation}
\label{eq:proj-orth}
\mathbf{W}_{k}^{T} \cdot \mathbf{W}_{k} = (\mathbf{I} - 2 \cdot \mathbf{r}_{k} \cdot \mathbf{r}_{k}^{T}) \cdot (\mathbf{I} - 2 \cdot \mathbf{r}_{k} \cdot \mathbf{r}_{k}^{T}) = \mathbf{I}
\end{equation}
This indicates that the projection operation corresponds to a rotation operation specific to relation $k$, which is proven to be highly beneficial for the EA task~\cite{rrea}.

In Eq.~(\ref{eq:sample}), we particularly add a term $\mathbf{e}_{i}^{T}  \cdot \mathbf{W}_{p} \cdot \mathbf{e}_{y}$ to reflect the influence of root entity $i$ on the selection of its second-order neighbor $y$. As shown in Fig.~\ref{fig:path}, although there is no direct relation between $i$ and $y$, we can compose a dummy relation $p$ along the path that connects them: $p = l \circ k$, where $\circ$ denotes the composition operator between relations $l=\phi(i,x)$ and $k=\phi(x,y)$. In practice, we use the Hadamard product to achieve this:
\begin{equation}
\label{eq:hadamard}
\mathbf{r}_{p} = \mathbf{r}_{l} \odot \mathbf{r}_{k}
\end{equation}
Then, the projection matrix is accordingly set as follows:
\begin{equation}
\label{eq:proj-w-p}
\mathbf{W}_{p} = \mathbf{I} - 2 \cdot \mathbf{r}_{p} \cdot \mathbf{r}_{p}^{T}
\end{equation}
Note that, similar to the projection matrix $\mathbf{W}_{k}$, the projection matrix $\mathbf{W}_{p}$ is also orthogonal.

Through hierarchical sampling, we can customize a personalized tree neighborhood for each target entity $i$, which naturally serves as the root of the tree. For each sampled tree, its aggregation paths form a subset of the traditional GNN's aggregation paths. By repeating random sampling multiple times, our parametric sampling technique can gradually cover all the possible aggregation paths.

\subsection{Tree Attention Encoding}
Based on the sampled tree, we accordingly develop an innovative tree attention encoder to obtain the embedding of the root entity. The encoding process is similar to the reverse process of sampling, as illustrated by arrows in Fig.~\ref{fig:path}. We use different attention mechanisms for the sampling process and the encoding process because the former operates on the original KG for sampling while the latter operates on the sampled tree for encoding. Referring to Fig.~\ref{fig:path} again, the aggregation attention coefficient from entity $y$ to entity $x$ is computed as follows:
\begin{equation}
\label{eq:tree-attn-unnormal}
b^{(i)}_{x,k,y} = \sigma(\mathbf{w}_{att}^{T} \cdot [\mathbf{W}_{p}^{T} \cdot \mathbf{e}_{i} \| \mathbf{W}_{k}^{T} \cdot \mathbf{e}_{x} \| \mathbf{e}_{y}])
\end{equation}
\begin{equation}
\label{eq:tree-attn-normal}
a^{(i)}_{x,k,y} = \frac{ exp(b^{(i)}_{x,k,y}) }{ \sum\limits_{z \in \mathcal{C}^{(i)}_x, s = \phi(x,z)} exp(b^{(i)}_{x,s,z}) }
\end{equation}
where $\mathbf{w}_{att}$ is the attention parameter, and $\mathcal{C}^{(i)}_x$ denotes the children of entity $x$ in the sampled rooted tree. Note that in Eq.~(\ref{eq:sample}), we use $\mathbf{W}_{p}$ and $\mathbf{W}_{k}$ to project $\mathbf{e}_{y}$ to $\mathbf{e}_{i}$ and $\mathbf{e}_{x}$. Here, in Eq.~(\ref{eq:tree-attn-unnormal}), we use $\mathbf{W}_{p}^{T}$ and $\mathbf{W}_{k}^{T}$ to project $\mathbf{e}_{i}$ and $\mathbf{e}_{x}$ back to $\mathbf{e}_{y}$. This is because according to the orthogonality property, we have: $\mathbf{W}_{p}^{-1} = \mathbf{W}_{p}^{T}$ and $\mathbf{W}_{k}^{-1} = \mathbf{W}_{k}^{T}$. Then, the entity embedding of entity $x$ is updated as follows:
\begin{equation}
\label{eq:attn-agg}
\mathbf{e}_x' = \sigma( \mathbf{e}_{x} + \sum\limits_{y \in \mathcal{C}^{(i)}_x, k = \phi(x,y)} a^{(i)}_{x,k,y} \cdot \mathbf{W}_{k}^{T} \cdot \mathbf{e}_{y} )
\end{equation}
By hierarchical aggregation from the descendants towards the root, we can finally obtain the embedding of the root entity $i$.

\subsection{Generating Pseudo Labels for Alignment}
Based on the obtained entity embeddings, we can compute a matrix $\mathbf{S}$ to describe the similarities between each pair of entities that are from different KGs. For example, given a pair of entities $<i_1,j_2>$ from $\mathcal{G}_1$ and $\mathcal{G}_2$, respectively, we can compute their cosine similarity as follows:
\begin{equation}
\label{eq:cosine}
\mathbf{S}_{{i_1},{j_2}} = \frac{\mathbf{e}_{i_1} \cdot \mathbf{e}_{j_2}}{\|\mathbf{e}_{i_1} \| \|\mathbf{e}_{j_2}\|}
\end{equation}
In addition, we adopt the cross-domain similarity local scaling (CSLS) technique~\cite{csls} to mitigate the negative effects of the hubness of some high-degree entities. It re-processed the similarity matrix as follows:
\begin{equation}
\label{eq:csls}
{\widetilde{\mathbf S}}_{{i_1},{j_2}} = 2 \cdot \mathbf{S}_{{i_1},{j_2}} - \frac{1}{\delta} \cdot \sum\limits_{q \in \mathcal{N}_{i_1}^{(2)}}\mathbf{S}_{{i_1},{q}} - \frac{1}{\delta} \cdot \sum\limits_{p \in \mathcal{N}_{j_2}^{(1)}}{\mathbf{S}_{{p},{j_2}}}
\end{equation}
where $\mathcal{N}_{i_1}^{(2)}$ denotes the set of the entity $i$'s top-$\delta$ nearest neighbors in $\mathcal{G}_2$, and $\mathcal{N}_{j_2}^{(1)}$ has a similar meaning. Thus, the similarities between hub entities and other entities are decreased, giving fair consideration to isolated entities~\cite{openea, csls}.

The pseudo labels are automatically based on ${\widetilde{\mathbf S}}$, through a bi-directional matching process. That is, if two entities are the closest counterpart for each other between $\mathcal{G}_1$ and $\mathcal{G}_2$, then they are treated as a pseudo label for alignment, and will be added to a set $\mathcal{S}$, formally described as follows:
\begin{equation}
\label{eq:label}
\mathcal{S} = \mathcal{S} \cup \{<i_1,j_2> | {\widetilde{\mathbf S}}_{{i_1},{j_2}} \ge {\widetilde{\mathbf S}}_{{i_1},{:}} , {\widetilde{\mathbf S}}_{{i_1},{j_2}} \ge {\widetilde{\mathbf S}}_{{:},{j_2}}\}
\end{equation}
where ${\widetilde{\mathbf S}}_{{i_1},{:}}$ and ${\widetilde{\mathbf S}}_{{:},{j_2}}$ represent the elements of the ${i_1}$-th row and the ${j_2}$-th column of the matrix, respectively. We use the pseudo labels contained in $\mathcal{S}$ to guide the alignment training, and $\mathcal{S}$ is dynamically updated in every epoch.

\subsection{Contrastive Alignment Loss}
\label{subsec:align}
Now, we are ready to define the alignment loss. We adopt the InfoNCE loss~\cite{opengcl} to achieve the alignment task in a contrastive manner. First, given a pseudo-label denoted by an entity pair $<i_1,j_2>$, their similarity is measured by a function as follows:
\begin{equation}
\label{eq:distance}
g(i_1,j_2)= exp( \frac{\mathbf{e}_{i_1}^{T} \cdot \mathbf{e}_{j_2}}{\tau} )
\end{equation}
where $\tau$ denotes a temperature hyper-parameter. Then, the similarity of the pseudo-label is maximized while the similarity of randomly sampled entity pairs is minimized.
\begin{equation}
\label{eq:left-nce}
\overrightarrow{l} (i_1,j_2) = -log \big( \frac { g(i_1,j_2) } { g(i_1,j_2) + \sum\limits_{y_2 \neq j_2, y_2 \sim P_{\mathcal{E}_2}} g(i_1,y_2)} \big)
\end{equation}
\begin{equation}
\label{eq:right-nce}
\overleftarrow{l} (i_1,j_2) = -log \big( \frac { g(i_1,j_2) } { g(i_1,j_2) + \sum\limits_{x_1 \neq i_1, x_1 \sim P_{\mathcal{E}_1}} g(x_1,j_2)} \big)
\end{equation}
For each pseudo-label $<i_1,j_2>$, we generate negative examples from both directions. One sub-loss fixes $i_1$ and randomly samples entities from $\mathcal{E}_2$, i.e., by distribution $P_{\mathcal{E}_2}$. The other sub-loss fixes $j_2$ and randomly samples entities from $\mathcal{E}_1$, i.e., by distribution $P_{\mathcal{E}_1}$. The final contrastive alignment loss is computed by averaging the two sub-losses:
\begin{equation}
\label{eq:loss-align}
\mathcal{L}_A = \frac{1}{2 \cdot |\mathcal{S}|} \sum_{<i_1,j_2> \in \mathcal{S}} \overrightarrow{l} (i_1,j_2) + \overleftarrow{l} (i_1,j_2)
\end{equation}
where $\mathcal{S}$ is the set of pseudo alignment labels generated by Eq.~(\ref{eq:label}). By definition, $\mathcal{L}_A$ maximizes the mutual information between entity pairs in $\mathcal{S}$.

\subsection{Mutual Information-based Regularization}
As described by Eq.~(\ref{eq:labse}), we initialize the entity embeddings by inputting their names into LLMs. The previous work~\cite{selfkg} has demonstrated that quite good EA performance can be achieved by using only these initialized entity embeddings, which can serve as a ``weak supervision signal'' for the EA task. Inspired by this finding, we develop a correction mechanism to address Limitation 2 mentioned in Section~\ref{sec:introduction}. The basic idea is to prevent the encoder from losing the beneficial information of these initial entity embeddings. To this end, we introduce three regularization terms based on mutual information, as follows:
\begin{equation}
\label{eq:reg-entity}
l_{\mathcal{E}} = \text{InfoNCE}(\mathbf{e}_i, \mathbf{f}_i^e)
\end{equation}
\begin{equation}
\label{eq:reg-relation}
l_{\mathcal{R}} = \text{InfoNCE}(\mathbf{r}_k, \mathbf{f}_k^r)
\end{equation}
\begin{equation}
\label{eq:reg-triple}
l_{\mathcal{T}} = \text{CE}(w_{i,j}, \mathbf{A}_{i,j})
\end{equation}
As shown in Eqs.~(\ref{eq:reg-entity}, \ref{eq:reg-relation}), we maximize the mutual information between entity/relation embeddings and their name features by minimizing the InforNCE loss as described by Eqs.~(\ref{eq:distance}, \ref{eq:left-nce}, \ref{eq:right-nce}) above. For Eq.~(\ref{eq:reg-triple}), following previous work~\cite{gmi}, we minimize the cross-entropy loss to maximize the mutual information between the estimated edge weight $w_{i,j}$ and the input edge weight $\mathbf{A}_{i,j}$, and the former is computed as follows:
\begin{equation}
w_{i,j} = \sigma(\mathbf{e}_i^{T} \cdot \mathbf{W}_n \cdot \mathbf{e}_j) 
\end{equation}
where $\mathbf{W}_n$ is a parameter matrix. 

These three mutual information-based terms can help the model preserve the information of the initialized entity features, the initialized relation features, and the input KG structural features, respectively. The total regularization loss is computed as the sum of these three terms:
\begin{equation}
\label{eq:mi-reg}
\mathcal{L}_{MI} = l_{\mathcal{E}} + l_{\mathcal{R}} + l_{\mathcal{T}}
\end{equation}

\subsection{Training}
\label{subsec:train}
Finally, the overall loss is a weighted combination of the alignment loss and the regularization loss:
\begin{equation}
\label{eq:loss-overall}
\mathcal{L} = \lambda \cdot \mathcal{L}_A + (1 - \lambda) \cdot\mathcal{L}_{MI}
\end{equation}
where $\lambda$ is a balance hyper-parameter. All the model parameters are optimized under the guidance of the overall loss. As the parameter optimization progresses, the model will gradually learn better embeddings, sample more personalized trees, learn more discriminative aggregation paths, and generate more effective pseudo-labels, finally leading to better EA performance.

After the model training is finished, the distance between entity embeddings can reflect the semantic similarity between these entities. Thus, we can discover more potentially aligned entity pairs across different KGs by measuring their embedding distance.

Recall that in Section~\ref{sbusec:sample-tree}, we sample personalize tree neighborhood according to the attention distribution, which depends on the embeddings of entities and relations, as described by Eqs.~(\ref{eq:sample-att}-\ref{eq:proj-w-p}). However, as a discrete operator, the sampling operation is not differentiable. To address this issue, we adopt the momentum technique~\cite{moco} to update the entity and relation embeddings for sampling. For higher efficiency, we sample trees and update pseudo-labels every $m$ epoch.

Algorithm~\ref{alg:unea} shows the overall training procedure of UNEA.

\begin{algorithm}[ht]
\caption{The training procedure of UNEA.}
\label{alg:unea}
\textbf{Input}: Two candidate KGs to be aligned $\mathcal{G}_1$ and $\mathcal{G}_2$.\\
\textbf{Output}: Entity embeddings.
\begin{algorithmic}[1] 
\STATE Randomly initialize trainable model parameters;
\STATE Use LLM to initialize entity and relation embeddings according to Eqs.~(\ref{eq:labse}-\ref{eq:init});
\WHILE{not converge}
    \STATE Sample personalized rooted trees according to the attention distribution according to Eqs.~(\ref{eq:sample}-\ref{eq:proj-w-p});
    \STATE Perform tree attention encoding by Eqs.~(\ref{eq:tree-attn-unnormal}-\ref{eq:attn-agg});
    \STATE Generate pseudo labels according to Eqs.~(\ref{eq:cosine}-\ref{eq:label});
    \STATE Compute alignment sub-loss by Eqs.~(\ref{eq:distance}-\ref{eq:loss-align});
    \STATE Compute regularization sub-loss by Eqs.~(\ref{eq:reg-entity}-\ref{eq:mi-reg});
    \STATE Compute the overall loss by Eq.~(\ref{eq:loss-overall});
    \STATE Update model parameters by gradient descent;
\ENDWHILE
\end{algorithmic}
\end{algorithm}

\section{Experiment}
\label{sec:experiment}
In this section, we conduct extensive experiments to show the superior effectiveness of our proposed UNEA.

\subsection{Datasets}
\label{subsec:datasets}
We use the DBP15K dataset~\cite{jape} and the DWY100K dataset~\cite{bootea} in our experiments, which are the most widely used benchmark datasets in previous studies. Table~\ref{dataset} summarizes the key statistics of the datasets.

\begin{table}[ht]
\centering
\tabcolsep=0.19cm
\caption{Dataset statistics.}
\label{dataset}
\renewcommand{\arraystretch}{1.3}
\begin{tabular}{|c|c|c|c|c|}
\hline
\textbf{Datasets} & \textbf{KGs} & \textbf{\# Entities} & \textbf{\# Relations} & \textbf{\# Triples} \\
\hline
\multicolumn{5}{|c|}{\textbf{DBP15K}} \\
\hline
\multirow{2}{*}{DBP15K$_{\text{zh\_en}}$} & Chinese &19388& 1700&70414 \\
& English &19572 &1322 & 95142 \\ 
\hline
\multirow{2}{*}{DBP15K$_{\text{ja\_en}}$} & Japanese & 19814 & 1298 & 77214 \\
& English &19780 & 1152 & 93484 \\
\hline
\multirow{2}{*}{DBP15K$_{\text{fr\_en}}$} & French & 19661 &902 & 105998 \\ 
& English & 19993 & 1207 & 11572 \\
\hline
\multicolumn{5}{|c|}{\textbf{DWY100K}} \\
\hline
\multirow{2}{*}{DWY100K$_{\text{dbp\_wd}}$} & DBpedia & 100000 & 330 & 463294 \\ 
& Wikipedia & 99990 & 220 & 448736 \\ 
\hline
\multirow{2}{*}{DWY100K$_{\text{dbp\_yg}}$} & DBpedia & 100000 & 302 & 428952 \\ 
& YAGO3 & 100000 & 31  & 502563 \\ 
\hline
\end{tabular}
\end{table}

\begin{itemize}
\item \textbf{DBP15K} is extracted from three language versions of DBpedia. It consists of three subsets for alignment: Chinese-English (zh-en), Japanese-English (ja-en), and French-English (fr-en), each containing 15K aligned entity pairs.
\item \textbf{DWY100K} is extracted from DBpedia, Wikidata, and YAGO. It consists of two subsets for alignment: DBpedia-Wikidata (dbp-wd) and DBpedia-YAGO (dbp-yg), each containing 100K aligned entity pairs. Considering that the DWY100K dataset represents entities by their indices, following previous work~\cite{selfkg}, we extract their entity names by the Wikidata API\footnote{\url{https://pypi.org/project/Wikidata/}}.
\end{itemize}

\subsection{Baselines}
\label{subsec:baselines}
We compare our UNEA against eighteen state-of-the-art baseline EA methods, which can be divided into three categories as follows:
\begin{itemize}
\item \textbf{Supervised \& Trans-based methods}: MTransE~\cite{mtranse}, TranseEdge~\cite{transedge}, JAPE~\cite{jape}, BootEA~\cite{bootea}, and MRPEA~\cite{mrpea}.
\item \textbf{Supervised \& GNN-based methods}: GCN-Align~\cite{gcn-align}, MuGNN~\cite{mugnn}, AliNet~\cite{alinet}, RDGCN~\cite{rdgcn}, HGCN~\cite{hgcn-je-jr}, RNM~\cite{rnm}, NAEA~\cite{naea}, and RREA~\cite{rrea}.
\item \textbf{Unsupervised methods}: MultiKE~\cite{multike}, EVA~\cite{eva}, SelfKG~\cite{selfkg}, ICLEA~\cite{iclea}, and UPLR~\cite{uplr}.
\end{itemize}


\begin{table*}[ht]
\centering
\tabcolsep=0.45cm
\renewcommand{\arraystretch}{1.3}
\caption{Alignment performance comparison on DBP15K.}
\label{tab:accuracy-dbp}
\begin{tabular}{|c|c|c|c|c|c|c|c|c|c|}
\hline
\multirow{2}[0]{*}{Models} & \multicolumn{3}{c|}{DBP15K$_{\text{zh\_en}}$} & \multicolumn{3}{c|}{DBP15K$_{\text{ja\_en}}$} & \multicolumn{3}{c|}{DBP15K$_{\text{fr\_en}}$} \\
\cline{2-10}
& Hits@1 & Hits@10 &MRR & Hits@1 & Hits@10 &MRR & Hits@1 & Hits@10 &MRR \\ 
\hline
\multicolumn{10}{|c|}{\textbf{Supervised \& Trans-based}} \\
\hline
MTransE &  0.308 & 0.614 & 0.364 & 0.279 & 0.575& 0.349 & 0.244& 0.556 & 0.335  \\ 
\hline
JAPE &  0.412 & 0.745 & 0.490 & 0.363 & 0.685&0.476 &  0.324& 0.667 & 0.430  \\ 
\hline
BootEA &  0.629 & 0.848 & 0.703 & 0.622 & 0.854 &0.701 & 0.653& 0.874 & 0.731  \\ 
\hline
MRPEA &  0.681 & 0.867 & 0.748 & 0.655 & 0.859 & 0.727& 0.677& 0.890 & 0.755 \\ 
\hline
TransEdge &  0.735 & 0.919 & 0.801 & 0.719 & 0.932 & 0.795& 0.710& 0.941 & 0.796 \\ 
\hline
\multicolumn{10}{|c|}{\textbf{Supervised \& GNN-based}} \\
\hline
GCN-Align &  0.413 & 0.744  & 0.549& 0.399 & 0.745 & 0.546& 0.373& 0.745 & 0.532  \\ 
\hline
MuGNN &  0.494 & 0.844 & 0.611 & 0.501 & 0.857 & 0.621 & 0.495& 0.870 & 0.621 \\ 
\hline
Alinet &  0.679 & 0.785  &0.628  & 0.740 & 0.872  &0.645 & 0.894& 0.952  & 0.657 \\
\hline
RDGCN &  0.708 & 0.846  &0.746  & 0.767 & 0.895  & 0.812& 0.886& 0.957  & 0.911 \\ 
\hline
HGCN  &  0.720& 0.857  & 0.768 & 0.766 & 0.897  & 0.813& 0.933& 0.960  & 0.917 \\ 
\hline
RNM  &  0.840 & 0.919  &0.870  & 0.872 & 0.944  &0.899 & 0.938& 0.954  &0.954  \\ 
\hline
NAEA &  0.650 & 0.867  &0.720  & 0.641 & 0.873 & 0.718 & 0.673& 0.894  & 0.752 \\ 
\hline
RREA  &  0.801 & 0.938  & 0.857  & 0.802 & 0.952  & 0.858 & 0.827 &  0.966  &  0.881  \\ 
\hline
\multicolumn{10}{|c|}{\textbf{Unsupervised}} \\
\hline
MultiKE  &  0.509 & 0.576  & 0.544  & 0.393 & 0.498  & 0.439 & 0.639 & 0.712 & 0.696 \\ 
\hline
EVA & 0.761 & 0.907 & 0.814 & 0.762 & 0.913 & 0.817 & 0.793 & 0.942 & 0.847 \\
\hline
SelfKG &  0.773 & 0.855  & 0.824 & 0.806 & 0.882  & 0.847 & 0.931& 0.972  & 0.937 \\ 
\hline
ICLEA  &  0.834& 0.894  &  0.879& 0.842 & 0.917 & 0.885& 0.936& 0.973  & 0.947 \\
\hline
UPLR  &  0.857& 0.913  & 0.897& 0.903 & 0.947 &0.913 & 0.953& 0.991 & 0.960 \\ 
\hline
\textbf{UNEA}& \textbf{0.906} & \textbf{0.939} & \textbf{0.913} & \textbf{0.911} & \textbf{0.953}  & \textbf{0.929}& \textbf{0.973}& \textbf{0.993}  &\textbf{0.967} \\
\hline
\end{tabular}
\end{table*}

\begin{table*}[!ht]
\centering
\tabcolsep=0.825cm
\renewcommand{\arraystretch}{1.3}
\caption{Alignment performance comparison on DWY100K.}
\label{tab:accuracy-dwy}
\begin{tabular}{|c|c|c|c|c|c|c|}
\hline
\multirow{2}[0]{*}{Models} & \multicolumn{3}{c|}{DWY100K$_{\text{dbp\_wd}}$} & \multicolumn{3}{c|}{DWY100K$_{\text{dbp\_yg}}$} \\
\cline{2-7}
& Hits@1 & Hits@10 & MRR & Hits@1 & Hits@10 & MRR \\ 
\hline
\multicolumn{7}{|c|}{\textbf{Supervised \& Trans-based}} \\
\hline
MTransE &  0.281 & 0.520 & 0.363 & 0.252 & 0.493 & 0.334  \\ 
\hline
IPTransE & 0.349 & 0.638 & 0.447 & 0.297 & 0.558 & 0.386  \\ 
\hline
JAPE &  0.318 & 0.589 & 0.411 & 0.236 & 0.484 & 0.320  \\ 
\hline
BootEA &  0.747 & 0.898 & 0.801 & 0.761 & 0.894 & 0.808  \\ 
\hline
TransEdge & 0.692 & 0.898 & 0.770 & 0.726 & 0.909 & 0.792 \\ 
\hline
\multicolumn{7}{|c|}{\textbf{Supervised \& GNN-based}} \\
\hline
GCN-Align & 0.506 & 0.772 & 0.600 & 0.597 & 0.838 & 0.682 \\ 
\hline
MuGNN & 0.616 & 0.897 & 0.714 & 0.741 & 0.937 & 0.810  \\ 
\hline
Alinet & 0.690 & 0.908 & 0.766 & 0.786 & 0.943 & 0.841  \\ 
\hline
NAEA & 0.767 & 0.918 & 0.817 & 0.779 & 0.913 & 0.821  \\ 
\hline
RREA & 0.854 & 0.966 & 0.877 & 0.874 & 0.976 & 0.913  \\ 
\hline
\multicolumn{7}{|c|}{\textbf{Unsupervised}} \\
\hline
MultiKE & 0.915 & 0.952 & 0.928 & 0.880 & 0.953 & 0.906 \\ 
\hline
SELKG & 0.983 & 0.998 & 0.968 & 0.997 & 1.000 & 0.965  \\ 
\hline
ICLEA & 0.985& 0.994 & 0.965 & 0.998 & 1.000 & 0.972 \\ 
\hline
UPLR & 0.988 & 0.996 & 0.973 & 0.998 & 1.000 & 0.977 \\ 
\hline
\textbf{UNEA} &  \textbf{0.991} & \textbf{0.998} & \textbf{0.977} & \textbf{1.000} & \textbf{1.000} & \textbf{0.986}  \\ 
\hline
\end{tabular}
\end{table*}

\subsection{Implementation Details}
\label{subsec:setting}
We implement our UNEA by PyTorch\footnote{\url{https://github.com/wzCSDN/UNEA}.}. All the trainable parameters are first randomly initialized by the Xavier distribution~\cite{xavier}. Then, the embeddings of entities and relations are initialized by Eqs.~(\ref{eq:labse},\ref{eq:init}), and we adopt a state-of-the-art multi-lingual pre-trained language model LaBSE~\cite{las-bert} as the LMM. Finally, the model parameters are optimized by gradient descent, and we adopt the Adam optimizer with a learning rate of 0.0001. The batch size is set to $128$. The number of epochs is set to 300, the embedding dimensionality of both entities and relations is set to 300, the temperature hyper-parameter $\tau$ is set to 0.08, the number of negative samples is set to $128$, the momentum is set to 0.9, and the non-linear activation function $\sigma(\cdot)$ is implemented as LeakyReLU. For supervised baselines, following convention, 30\% of pre-aligned entity pairs are treated as the labels for supervision. For unsupervised baselines as well as our UNEA, they are trained in an unsupervised manner without any manual labels. For fairness, all the methods only utilize the names of entities and relations without the preprocess of Google Translate. All the experiments are run on an NVIDIA GPU with 24GB memory.

For quantitative evaluation, consistent with most previous studies, we use three widely used metrics Hits@1, Hits@10, and MRR (mean reciprocal rank)~\cite{zeng2021comprehensive, zhao2020experimental,sun13benchmarking, zhang2022benchmark,sun2022revisiting,zhang2023semi}.

\subsection{Main Results}
\label{subsec:accuracy}
We compare the alignment accuracy of our UNEA against all the baseline methods. Table~\ref{tab:accuracy-dbp} and Table~\ref{tab:accuracy-dwy} show the results on DBP15K and DWY100k, respectively. As we can see, our UNEA achieves the best performance in all the cases, indicating its superiority. Besides, UNEA and several other unsupervised baselines can even outperform most of the supervised methods, which demonstrates the advancement of the unsupervised learning paradigm. Among supervised baselines, GNN-based methods achieve better overall performance than Trans-based methods, which is consistent with the findings of most of the previous studies. TransEdge shows the best performance compared with other Trans-based methods, which may be due to its effective bootstrapping strategy and context projection. RREA outperforms other GNN-based methods since it can satisfy two key criteria: (1) relational differentiation and (2) dimensional isometry, through orthogonalizing projection matrices. Among unsupervised baselines, ICLEA and UPLR outperform SelfKG because ICLEA additionally considers cross-KG interaction, and UPLR can iteratively bootstrap pseudo-labels. Our UNEA performs better than the other unsupervised baselines because it can learn more flexible embeddings by personalized tree neighborhood sampling and encoding, and mitigate distribution distortion by mutual information-based regularization. It's noteworthy that our UNEA achieves a 100\% Hits@1 score and a 100\% Hits@10 score on DWY100K$_{\text{dbp\_yg}}$, showcasing its powerful effectiveness for the EA task.

\subsection{Case Study} 
\label{subsec:case}

\begin{figure}[ht]
\centering
\includegraphics[width=1.0\columnwidth]{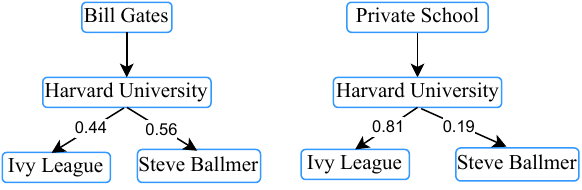}
\caption{Visualization of the sampled trees rooted at two entities. The aggregation weights are marked on the branches.}
\label{fig:case_1}
\end{figure}

We further analyze the effectiveness of UNEA more intuitively. In Fig.~\ref{fig:case_1}, we visualize the sampled trees of two target entities ``Bill Gates'' and ``Private School''. The attention coefficients for aggregation are also accordingly marked on the branches. We can see, in the left tree, ``Bill Gates'' entity assigns a larger coefficient to its grandson entity ``Steve Ballmer''. Contrastively, in the right tree, ``Private School'' entity assigns a larger coefficient to its grandson entity ``Ivy League''. This is more reasonable and intuitive in practice. Traditional GNN-like encoders cannot well capture this flexibility since there are always some sub-paths that are shared among different target entities.

\begin{figure}[!ht]
\centering
\includegraphics[width=1.0\columnwidth]{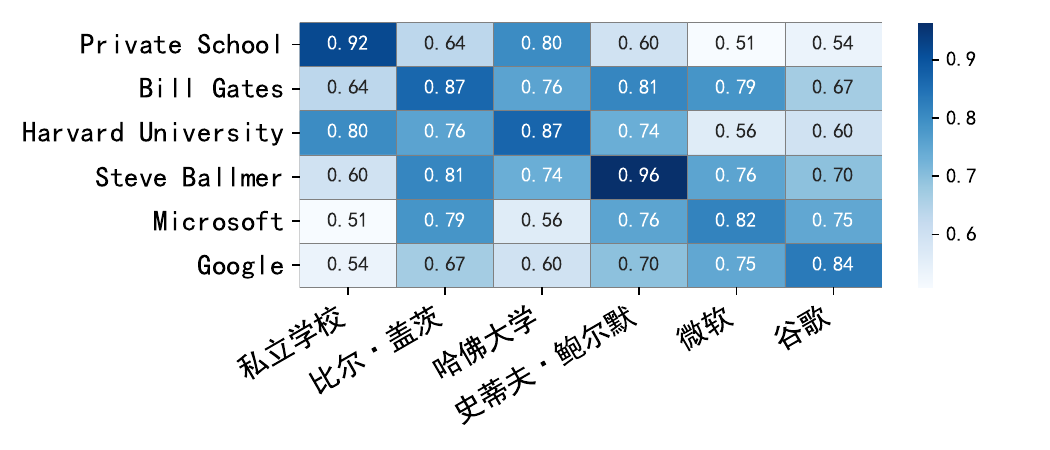}
\caption{Visualization of the embedding similarity between six pairs of entities. The values are re-normalized to the range $\left[0, 1\right]$.}
\label{fig:case_2}
\end{figure}

In Fig.~\ref{fig:case_2}, we visualize the embedding similarity between the six pairs of entities that are previously illustrated in Fig.~\ref{fig:exam_1}. The horizontal and vertical axes represent Chinese and English entities, respectively. As we can see, the diagonal values are significantly larger than others, which indicates that these pairs of entities can be properly aligned. In addition, we have the following observations: (1) ``Bill Gates'' is more similar to ``Microsoft'' than ``Google''; (2) ``Bill Gates'' is very similar to ``Steve Ballmer''; (3) ``Private School'' is very similar to ``Harvard University''. All three observations are highly consistent with real-world facts.

\subsection{Ablation Study} 
\label{sec:ablation}

\begin{table*}[ht]
\centering
\tabcolsep=0.46cm
\renewcommand{\arraystretch}{1.3}
\caption{Alignment performance comparison among different variants of UNEA.}
\label{tab:ablation}
\begin{tabular}{|c|c|c|c|c|c|c|c|c|c|}
\hline
\multirow{2}[0]{*}{Variants} & \multicolumn{3}{c|}{DBP15K$_{\text{zh\_en}}$} & \multicolumn{3}{c|}{DBP15K$_{\text{ja\_en}}$} & \multicolumn{3}{c|}{DBP15K$_{\text{fr\_en}}$} \\
\cline{2-10}
& Hits@1 & Hits@10 &MRR & Hits@1 & Hits@10 &MRR & Hits@1 & Hits@10 &MRR \\ 
\hline
\hline
UNEA-PS & 0.778 & 0.862  & 0.835 & 0.839 & 0.884  & 0.868& 0.911 & 0.936  & 0.930  \\
\hline
UNEA-TA & 0.844 & 0.900  & 0.883 & 0.862 & 0.924  & 0.897 & 0.940 & 0.966 & 0.950  \\ 
\hline
UNEA-MI & 0.851 & 0.904  & 0.886 & 0.893 & 0.927  & 0.895 & 0.944 & 0.979 & 0.952  \\ 
\hline
\hline
\textbf{UNEA}& \textbf{0.886} & \textbf{0.937} & \textbf{0.913} & \textbf{0.911} & \textbf{0.953}  & \textbf{0.929}& \textbf{0.973}& \textbf{0.993}  &\textbf{0.967} \\
\hline
\end{tabular}
\end{table*}

To verify the key components of our UNEA, we set three variants as follows: 
\begin{itemize}
\item \textbf{UNEA-PS} replaces the parametric sampling with random sampling based on uniform distribution;
\item \textbf{UNEA-TA} replaces the tree attention encoder with traditional GNN encoder~\cite{selfkg};
\item \textbf{UNEA-MI} removes the mutual information regularization (MI) module.
\end{itemize}


As shown in Table~\ref{tab:ablation}, all three variants perform worse than UNEA, indicating the effectiveness of the corresponding modules. Specifically, the variant UNEA-PS shows the worst accuracy because random sampling would introduce noisy neighbors that are not helpful for target entities. This is also consistent with the observation of~\cite{selfkg}. UNEA-PS shows the second-worst performance, suggesting that it is beneficial to learn personalized aggregation paths for target entities. UNEA performs better than UNEA-MI, which implies that mutual information-based regularization is helpful for the unsupervised EA task.

\begin{table*}[ht]
\centering
\tabcolsep=0.355cm
\renewcommand{\arraystretch}{1.3}
\caption{Comparison between traditional language model and large language model.}
\label{tab:init}
\begin{tabular}{|c|c|c|c|c|c|c|c|c|c|}
\hline
\multirow{2}[0]{*}{Model} & \multicolumn{3}{c|}{DBP15K$_{\text{zh\_en}}$} & \multicolumn{3}{c|}{DBP15K$_{\text{ja\_en}}$} & \multicolumn{3}{c|}{DBP15K$_{\text{fr\_en}}$} \\
\cline{2-10}
& Hits@1 & Hits@10 & MRR & Hits@1 & Hits@10 & MRR & Hits@1 & Hits@10 & MRR \\ 
\hline
\hline
only fastText & 0.631 & 0.746 & 0.692 & 0.703 & 0.795 & 0.739 & 0.852 & 0.907 & 0.880 \\
\hline
UNEA-MI (fastText) & 0.821 & 0.901 & 0.872 & 0.857 & 0.911 & 0.882 & 0.905 & 0.953 & 0.937 \\
\hline
UNEA+MI (fastText) & 0.844 & 0.924 & 0.890 & 0.873 & 0.925 & 0.903 & 0.933 & 0.968 & 0.950 \\
\hline
\textbf{Improvement (\%)} & \textbf{+2.80} $\uparrow$ & \textbf{+2.55} $\uparrow$ & \textbf{+2.06} $\uparrow$ & \textbf{+1.87} $\uparrow$ & \textbf{+1.54} $\uparrow$ & \textbf{+2.38} $\uparrow$ & \textbf{+3.09} $\uparrow$ & \textbf{+1.57} $\uparrow$ & \textbf{+1.39} $\uparrow$ \\
\hline
\hline
only LaBSE & 0.671 & 0.780 & 0.722 & 0.743 & 0.851 & 0.782 & 0.912 & 0.970 & 0.933 \\
\hline
UNEA-MI (LaBSE) & 0.851 & 0.904 & 0.886 & 0.893 & 0.927 & 0.895 & 0.944 & 0.979 & 0.952 \\
\hline
UNEA+MI (LaBSE) & 0.886 & 0.937 & 0.913 & 0.911 & 0.953 & 0.929 & 0.973 & 0.993 & 0.967 \\
\hline
\textbf{Improvement (\%)} & \textbf{+4.11} $\uparrow$ & \textbf{+3.65} $\uparrow$ & \textbf{+3.05} $\uparrow$ & \textbf{+1.98} $\uparrow$ & \textbf{+2.80} $\uparrow$ & \textbf{+3.80} $\uparrow$ & \textbf{+3.72} $\uparrow$ & \textbf{+1.43} $\uparrow$ & \textbf{+1.58} $\uparrow$ \\
\hline
\end{tabular}
\end{table*}

\subsection{Effectiveness of LLMs}
\label{sec:init}
While we utilize the LaBSE~\cite{las-bert} as the LLM to initialize the entity embeddings, some previous EA methods~\cite{eva,uea} adopt the traditional shallow language model fastText~\cite{fasttext} language model to initialize their entity embeddings. Here, we compare the impact of the two language models of LaBSE and fastText on the EA performance. The experimental results are shown in Table~\ref{tab:init}. 

We can see that LaBSE and its variants significantly outperform fastText and its variants in all cases, confirming that LaBSE is a stronger language model for EA, which is also consistent with the finding of SelfKG~\cite{selfkg}.
Besides, the two ``+MI'' UNEA variants outperform the two ``-MI'' UNEA variants, indicating that our MI module can effectively make use of the information contained in any pre-trained language model.
Finally, we compute the performance improvement brought by our MI module. As we can see, the improvement based on LaBSE is significantly greater than the improvement based on fastText, which implies that our MI module can amplify the superiority of LaBSE. This finding verifies our intuition: when the initialization quality is higher, the ``weak supervision signal'' is stronger, and the improvement brought by the regularization module is greater.

\begin{figure}[!ht]
\centering
\subfigure[$\lambda$ on Hits@1]{\includegraphics[width=0.49\columnwidth]{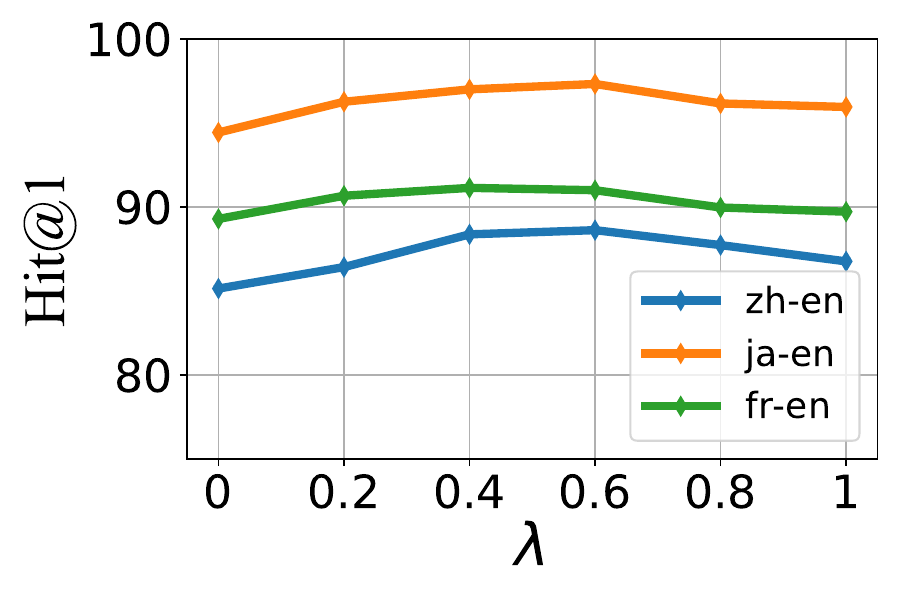}\label{fig:lambda_1}}
\subfigure[$\lambda$ on Hits@10]{\includegraphics[width=0.49\columnwidth]{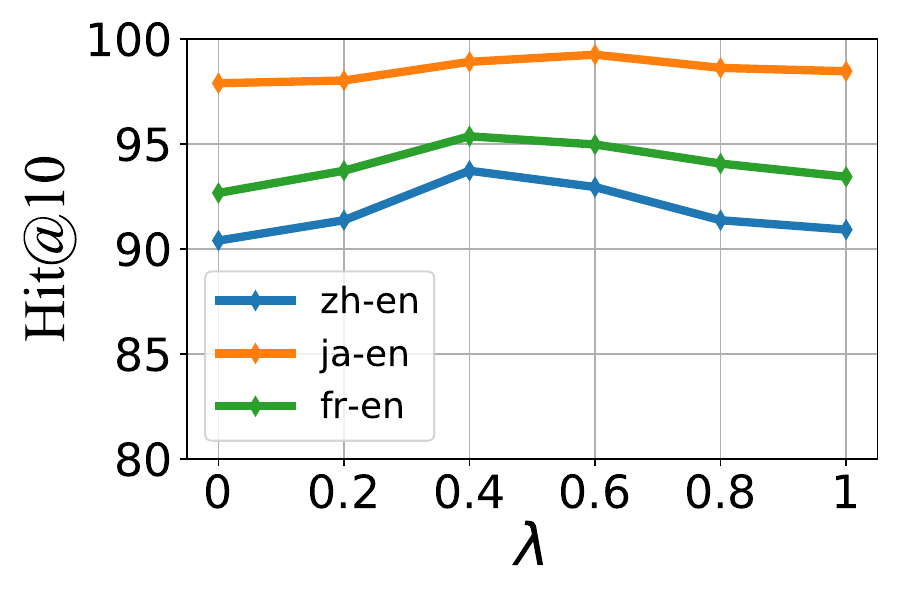}\label{fig:lambda_10}}\\
\subfigure[$m$ on Hits@1]{\includegraphics[width=0.49\columnwidth]{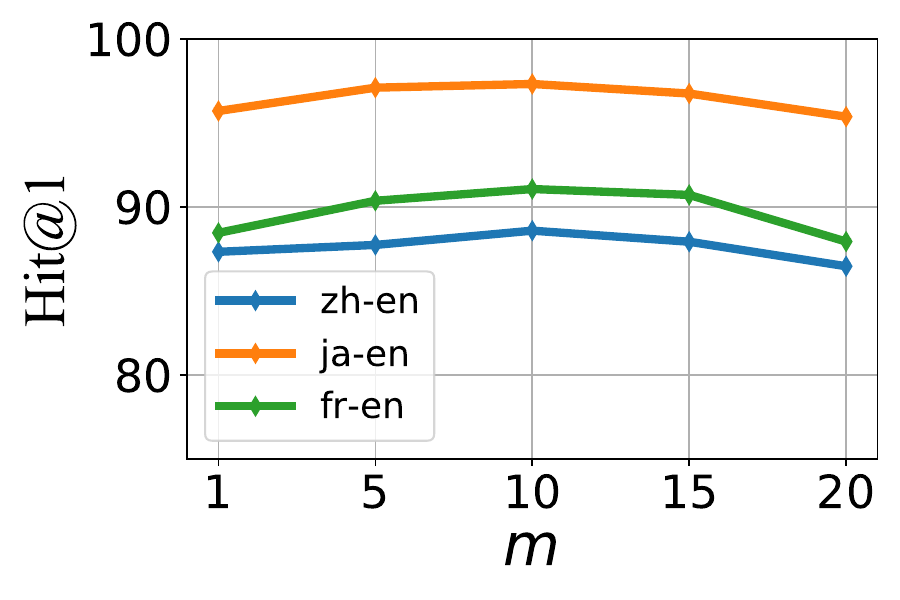}\label{fig:m_1}}
\subfigure[$m$ on Hits@10]{\includegraphics[width=0.49\columnwidth]{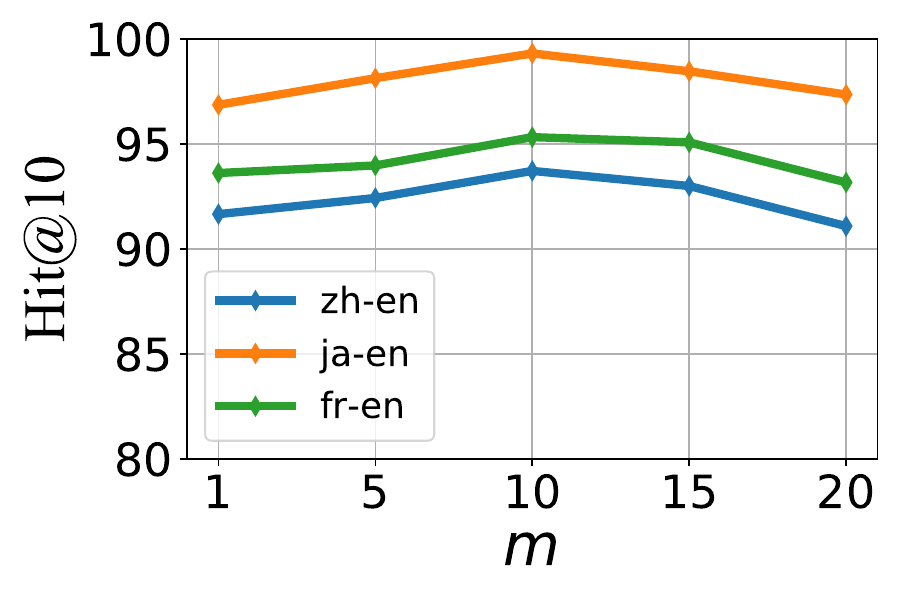}\label{fig:m_10}}
\caption{The sensitivity of balance hyper-parameter $\lambda$ and interval of epochs $m$.}
\label{fig:hyper-param}
\end{figure}

\subsection{Hyper-parameter Study} 
\label{subsec:para}
Referring to Section~\ref{subsec:train}, we have introduced two hyper-parameters. The hyper-parameter $\lambda$ balances the alignment loss and the mutual information-based regularization, and the hyper-parameter $m$ denotes the interval of the epochs for tree sampling and pseudo-label updating. Here, we investigate the effect of the two hyper-parameters on the EA performance.

As shown in Fig.~\ref{fig:hyper-param}, the two hyper-parameters are not very sensitive to EA performance. In Fig.~\ref{fig:lambda_1} and Fig.~\ref{fig:lambda_10}, when $\lambda = 0$, i.e., the regularization module is zeroed out, the performance is relatively low. When $\lambda \approx 0.4$, the model achieves its best performance. These phenomena verify that our regularization module is beneficial for the unsupervised EA task. In Fig.~\ref{fig:m_1} and Fig.~\ref{fig:m_10}, the performance is low when $m$ is too small or too large. This may be because the model cannot fully exploit pseudo-labels or cannot update pseudo-labels in time. In practice, we default $m$ to 10 for all the other experiments.

\section{Conclusion and Discussion}
\label{sec:conclu}
In this work, we note two limitations of existing EA methods. (1) They cannot flexibly capture the personality of entity embeddings due to the shared aggregation subpaths in their encoding procedures; (2) They cannot fully alleviate the distortion of the distribution similarity between candidate KGs in the unsupervised setting. To this end, we propose a novel unsupervised entity alignment method named UNEA. It samples a personalized tree neighborhood rooted at each target entity and learns personalized aggregation paths for the root entity. Three types of mutual information maximization-based regularization terms are introduced into the model to prevent the distribution distortion issue. Extensive experiments show that our UNEA achieves a new state-of-the-art performance in the EA task without any supervision information. It can even outperform previous supervised EA methods.

Although showing promising performance, our UNEA still has some limitations as follows. Like most unsupervised EA methods~\cite{selfkg,iclea,uplr}, UNEA requires unsupervised pre-trained language models to initialize the entity embeddings according to entities' names. Fortunately, in real life, the names of entities are usually available. The overall time complexity of our UNEA can reach the square level of the number of entities,   due to the bi-directional match of pseudo-labels. This complexity is on par with most previous unsupervised EA methods~\cite{uplr,iclea}.

\section*{Acknowledgments}
This work was supported in part by the National Natural Science Foundation of China under Grants 62303366, 62133012, 61936006, and 62073255, in part by the Innovation Capability Support Program of Shaanxi under Grant 2021TD-05, and in part by the Fundamental Research Funds for the Central Universities under Grants QTZX23039, XJSJ23030.

\bibliographystyle{IEEEtran}
\bibliography{unea}

\begin{IEEEbiography}[{\includegraphics[width=0.9in, height=1.25in, clip, keepaspectratio]{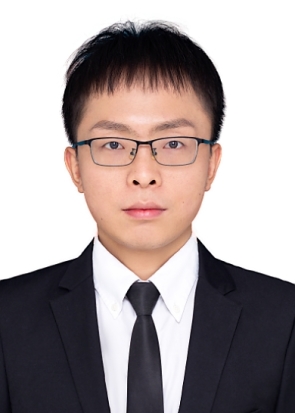}}]{Yaming Yang} received the B.S. and Ph.D. degrees in Computer Science and Technology from Xidian University, China, in 2015 and 2022, respectively. He is currently a lecturer with the School of Computer Science and Technology at Xidian University. His research interests include data mining and machine learning on graph data.
\end{IEEEbiography}

\begin{IEEEbiography}[{\includegraphics[width=0.9in, height=1.25in, clip, keepaspectratio]{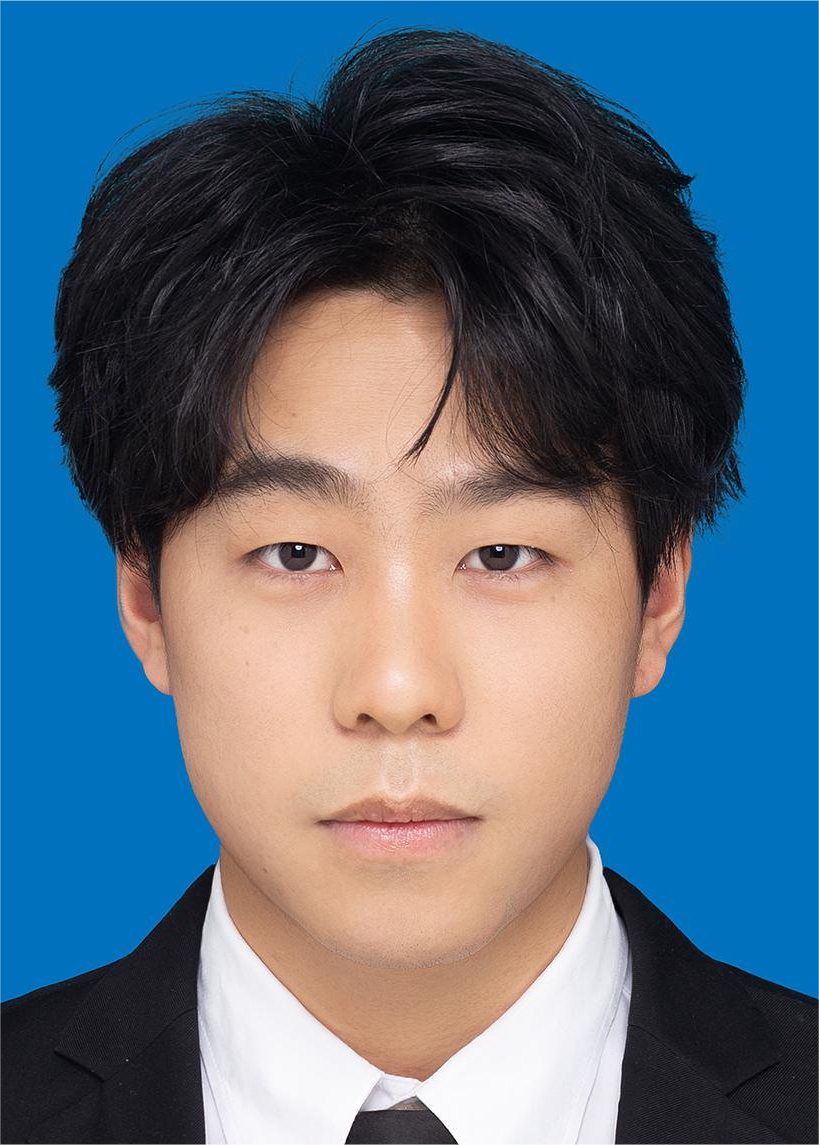}}]{Zhe Wang} received the B.S. degree in Management from Hefei University of Technology, China, in 2020. He is currently working towards a Ph.D. degree with the School of Computer Science and Technology, Xidian University, China. His research interests include data mining and machine learning on knowledge graph data.
\end{IEEEbiography}

\begin{IEEEbiography}[{\includegraphics[width=0.9in, height=1.25in, clip, keepaspectratio]{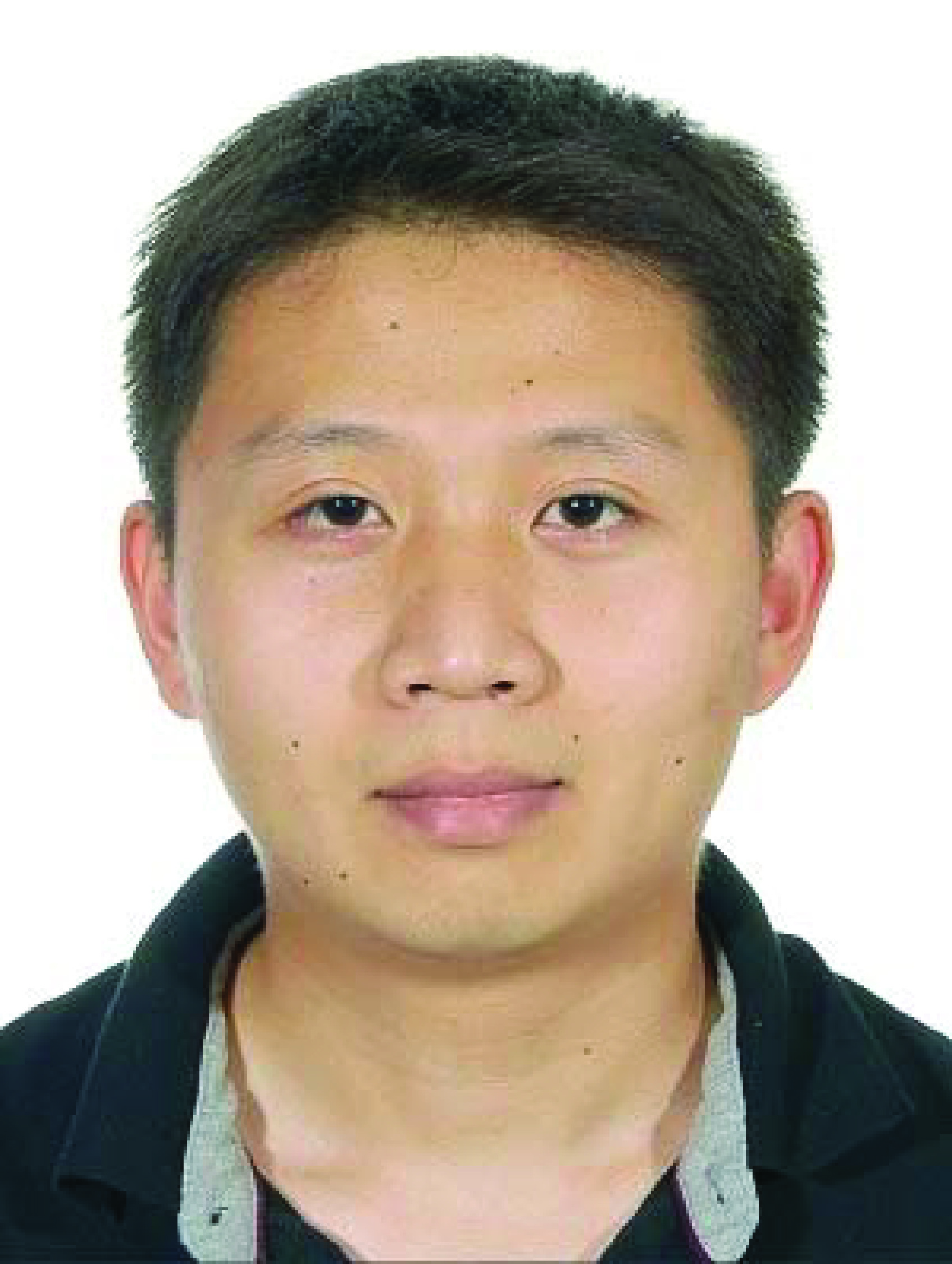}}]{Ziyu Guan} received the B.S. and Ph.D. degrees in Computer Science from Zhejiang University, Hangzhou China, in 2004 and 2010, respectively. He had worked as a research scientist in the University of California at Santa Barbara from 2010 to 2012, and as a professor in the School of Information and Technology of Northwest University, China from 2012 to 2018. He is currently a professor with the School of Computer Science and Technology, Xidian University. His research interests include attributed graph mining and search, machine learning, expertise modeling and retrieval, and recommender systems.
\end{IEEEbiography}

\begin{IEEEbiography}[{\includegraphics[width=0.9in, height=1.25in, clip, keepaspectratio]{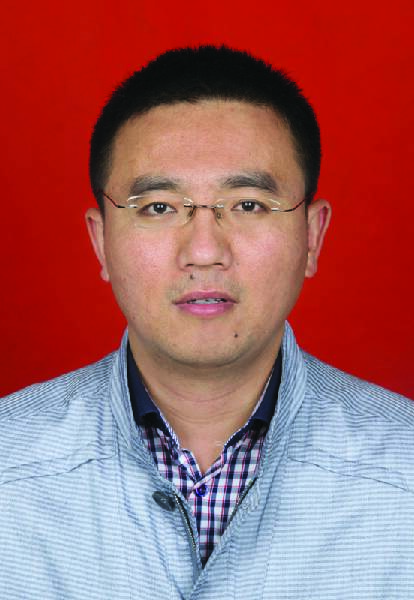}}]{Wei Zhao} received the B.S., M.S. and Ph.D. degrees from Xidian University, Xi’an, China, in 2002, 2005 and 2015, respectively. He is currently a professor in the School of Computer Science and Technology at Xidian University. His research direction is pattern recognition and intelligent systems, with specific interests in attributed graph mining and search, machine learning, signal processing and precision guiding technology.
\end{IEEEbiography}

\begin{IEEEbiography}[{\includegraphics[width=0.9in, height=1.25in, clip, keepaspectratio]{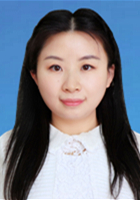}}]{Xinyan Huang} received the B.S. degree in Electronic Science and Technology from Shaanxi University of Science and Technology, Xi'an, China, in 2015, the M.S. degree in Electronic and Communication Engineering from Shaanxi Normal University, Xi'an, China, in 2017, and the Ph.D. degree in Computer Science and Technology from Xidian University, Xi'an, China, in 2024. She is currently working as a postdoctor at Xidian University. Her research interests include machine learning, computer vision, and pattern recognition.
\end{IEEEbiography}

\begin{IEEEbiography}[{\includegraphics[width=0.9in, height=1.25in, clip, keepaspectratio]{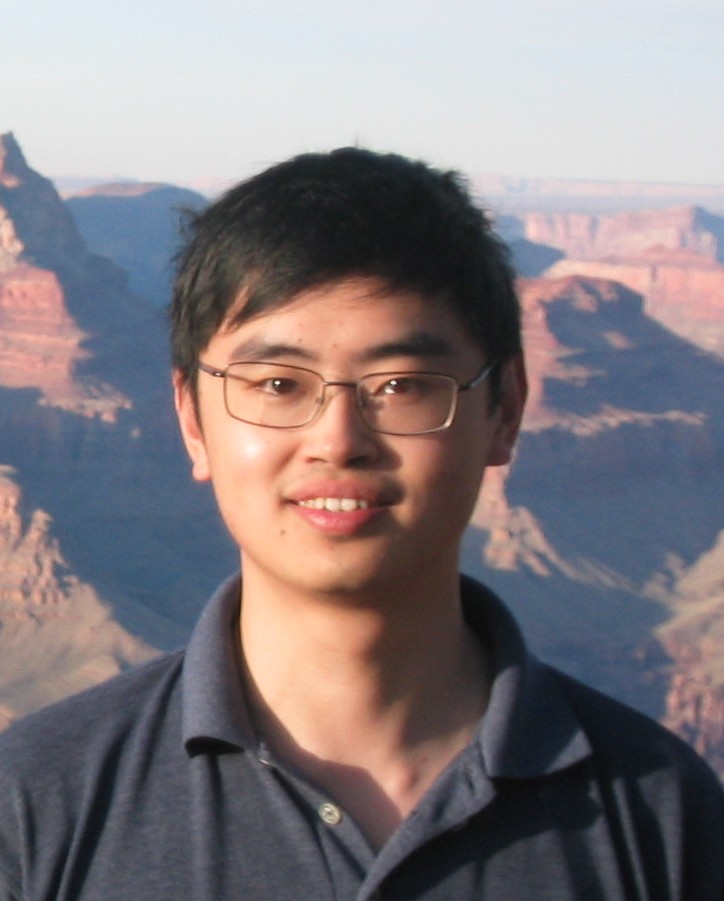}}]{Xiaofei He} received the B.S. degree in computer science from Zhejiang University, China, in 2000, and the Ph.D. degree in computer science from The University of Chicago, in 2005. He is currently a Professor with the State Key Lab of CAD\&CG, Zhejiang University. Prior to joining Zhejiang University, he was a Research Scientist with Yahoo! Research Labs, Burbank, CA, USA. His research interests include machine learning, information retrieval, and computer vision.
\end{IEEEbiography}

\vfill

\end{document}